# Plastic Contaminant Detection in Aerial Imagery of Cotton Fields with Deep Learning

Pappu Kumar Yadav[a*], J. Alex Thomasson[b], Robert G. Hardin[a], Stephen W. Searcy[a] ,Ulisses Braga-Neto[c], Sorin C. Popescu[d], Roberto Rodriguez[e], Daniel E Martin[f], Juan Enciso[a], Karem Meza[g], Emma L. White[a]

*Corresponding Author:  Pappu Kumar Yadav (pappuyadav@tamu.edu)

107, Price Hobgood Agricultural Engineering Research Lab

375 Olsen Blvd, College Station, TX 77840

[a] Department of Biological & Agricultural Engineering, Texas A&M University, College Station, TX

[b] Department of Agricultural & Biological Engineering, Mississippi State University, Mississippi State, MS

[c] Department of Electrical & Computer Engineering, Texas A&M University, College Station, TX

[d] Department of Ecology & Conservation Biology, Texas A&M University, College Station, TX

[e]U.S.D.A.- APHIS PPQ S&T, Insect Management and Molecular Diagnostics Laboratory, Edinburg, TX

[f] Aerial Application Technology Research, U.S.D.A. Agriculture Research Service, College Station, TX

[g] Department of Civil & Environmental Engineering, Utah State University, Logan, UT

## ABSTRACT

Plastic shopping bags that get carried away from the side of roads and tangled on cotton plants can end up at cotton gins if not removed before the harvest. Such bags may not only cause problem in the ginning process but might also get embodied in cotton fibers reducing its quality and marketable value. Therefore, it is required to detect, locate, and remove the bags before cotton is harvested. Manually detecting and locating these bags in cotton fields is labor intensive, time-consuming and a costly process. To solve these challenges, we present application of four variants of YOLOv5 (YOLOv5s, YOLOv5m, YOLOv5l and YOLOv5x) for detecting plastic shopping bags using Unmanned Aircraft Systems (UAS)-acquired RGB (Red, Green, and Blue) images. We also show fixed effect model tests of color of plastic bags as well as YOLOv5-variant on average precision (AP) , mean average precision (mAP@50) and accuracy. In addition, we also demonstrate the effect of height of plastic bags on the detection accuracy. It was found that color of bags had significant effect ($p < 0.001$) on accuracy across all the four variants while it did not show any significant effect on the AP with YOLOv5m ($p = 0.10$) and YOLOv5x ($p = 0.35$) at 95% confidence level. Similarly, YOLOv5-variant did not show any significant effect on the AP ($p = 0.11$) and accuracy ($p = 0.73$) of white bags, but it had significant effects on the AP ($p = 0.03$) and accuracy ($p = 0.02$) of brown bags including on the mAP@50 ($p = 0.01$) and inference speed ($p < 0.0001$). Additionally, height of plastic bags had significant effect ($p < 0.0001$) on overall detection accuracy. Using the desirability function, we found YOLOv5m to be the most desirable variant among all the four with desirability value of nearly 95% based on mAP@50, accuracies of white and brown bags, and inference speed. The findings reported in this paper can be useful in speeding up removal of plastic bags from cotton fields before harvest and thereby reducing the amount of contaminants that end up at cotton gins.

**1.0 Introduction**

Plastic contamination in U.S. cotton is a prevalent issue and a matter of concern for both ginners and the entire supply chain of the cotton industry. Considering the severity of the problem, the U.S. Department of Agriculture's Agricultural Marketing Service (USDA-AMS) began implementing new extraneous matter codes (71 and 72) for cotton classification in 2018 (Robbins, 2018). Plastic contaminants in cotton significantly reduce the fiber's marketable value, lowering the price paid by yarn spinners to growers (Himmelsbach et al., 2006; Pelletier et al., 2020; Whitelock et al., 2018). Plastic contaminants in cotton can come from various sources like plastic wraps on round cotton modules, plastic mulch used in crop production, and from other sources at various stages of supply chain starting in the field (Wanjura et al., 2020; Yadav et al., 2020). Contaminants may come from plastic rubbish, commonly plastic shopping bags, that are carried by wind from roadsides and then become entangled with cotton plants across cotton fields (Blake et al., 2020). During harvesting, mechanical cotton pickers and strippers do not separate this material from the cotton, and it becomes embedded in seed cotton modules. Plastic contaminants often find their way into the processing machinery of cotton gins (Pelletier et al., 2020) and become embedded in cotton bales that are ultimately shipped to spinners. Efforts are being made to design a machine-vision (MV) based automated control system at cotton gins to detect and remove plastic contaminants (Pelletier et al., 2020). However, detecting, locating, and removing plastic contaminants in the field before harvest can minimize the amount that ends up at the gin, thereby reducing the requirement for MV detection at that point. The only current option for field removal is human observation, which would be too time-consuming, costly, and inefficient to be a viable process. To solve the problem, earlier studies used an unmanned aircraft systems (UAS) with a multispectral camera to detect and locate shopping bags in cotton fields (Hardin et al., 2018; Yadav et al., 2020). A classifier based on a classical machine learning (ML) algorithm was trained with image-based spectral and textural features of plastic shopping bags (Yadav et al., 2020). The method resulted in a detection accuracy of over 64% in a cotton field before defoliation. Besides having inadequate detection accuracy – 90% or better is desirable – the method was time-consuming, requiring field data collection and offline image processing, meaning the plastic bags could only be located after a day or so, when images had been processed with the developed algorithm.

Traditional approaches of object detection (e.g., those in Hardin et al., 2018; and Yadav et al., 2020) involve feature extraction based on either histogram analysis – such as Histogram of Oriented Gradient (HOG) (Dalal & Triggs, 2005; Ren & Ramanan, 2013) and Edge Orientation Histograms (EOH) (Whitehill et al., 2009) – or on image textures (Kalinke et al., 1998). Both the HOG and EOH count occurrences of gradient orientation in a localized space of an image, while texture-based approaches calculate spatial variation of image tones depending upon statistics like contrast, entropy, etc. (Chen, 1995). Various methods have been tested to detect and identify foreign matter in cotton. For instance, Himmelsbach et al. (2006) used Fourier Transform Infrared spectroscopy to detect contaminants in cotton fibers. Based on differences in absorbance over a range of wavelengths, foreign particles in cotton fiber were identified. This kind of approach is practical for laboratory settings but not in the field, because detection with this method requires controlled conditions and is slow and expensive.

Deep learning (DL), on the other hand, can detect objects quickly in the field with good accuracy, desirable for efficient contamination removal in the field. To achieve near real-time detection and location of plastic bags in cotton fields, with improved detection accuracy, DL based on remotely sensed images is an attractive approach. DL-based algorithms have been successively used for object detection tasks in many agricultural applications (Chen et al., 2002; Fan et al., 2020; G. Liu et al., 2020;Yadav et al., 2022a;Yadav et al., 2022b;Yadav et al., 2022c). Among many DL-based algorithms, You Only Look Once version 5 (YOLOv5) is a state-of-the-art CNN architecture for real-time object detection. Similar to its earlier versions, YOLOv5 is a one-stage detector in which regression is used for detection and localization, but it is much faster and more accurate than previous versions, with inference speeds of up to 110 frames per second (FPS) (Brungel & Friedrich, 2021; Zhou et al., 2021) and 142 FPS (Solawetz, Jacob;Nelson, 2020) as compared to the older versions' average inference speed of 33 FPS (Mao et al., 2019; Redmon & Farhadi, 2018). Inference speed varies depending upon hardware, image quality, and size and number of objects present in the images. The end goal of this research is to deploy a trained model for real-time plastic bag detection, which requires high inference speeds, so YOLOv5 was selected for this study. Until the release of YOLOX in 2021 (Ge et al., 2021), YOLOv5 was the latest version of the YOLO family of object detection algorithms. It has four variants: YOLOv5s, YOLOv5m, YOLOv5l and YOLOv5x; where *s, m, l,* and *x* represent small, medium, large, and extra-large versions, respectively, in terms of network depth and number of parameters. The original *s, m, l,* and *x* versions have 283, 391, 499 and 607 layers and 7066239, 21060447, 46636735 and 87251103 parameters, respectively, while the scaled down trained models have 224, 308, 392  and 476 layers, and 7,056,607, 21,041,679, 46,605,951, and 872,105,423 parameters, respectively. In the current study, the four scaled down versions were used. The objectives of this study are as follows: (i) to determine how well YOLOv5 can be used to detect plastic shopping bags in a cotton field, (ii) to determine how the color of plastic shopping bags affects average precision (AP) and accuracy, (iii) to determine how YOLOv5-variant (*s, m, l* and *x*) affects AP, accuracy, mean average precision (mAP@50), and inference speed for different color bags, (iv) to identify an optimal YOLOv5-variant based on a desirability function dependent upon AP for brown and white bags, mAP@50, accuracy, and inference speed and (v) to determine the effect of height of plastic bags on cotton plants on the overall detection accuracy of the most desirable YOLOv5-variant.

## 2.0 Materials and Methods

### 2.1 Experiment Site

To mimic the natural occurrence of grocery-store plastic shopping bags in cotton fields, we manually tied a total of 180 plastic bags, 90 white and 90 brown, on cotton plants in a cotton field (26º 9' 51.62" N, 97º 56' 29.66" W) located near Weslaco in Hidalgo County, Texas (Fig. 1). The bags were tied at three different heights (top, middle, and bottom) on the plants, before defoliation. The order in which the bags were tied on plants followed a randomized sequence generated by Microsoft Excel software (Microsoft Corporation, Redmond, Washington, U.S.A.). The field had been planted with cotton seed of the Phytogen 350 W3FE variety (CORTEVA agriscience, Wilmington, Delaware). The soil at the experimental site includes two types: *Hidalgo sandy clay loam* and *Raymondville clay loam* (USDA-Natural Resources Conservation Service, 2020).

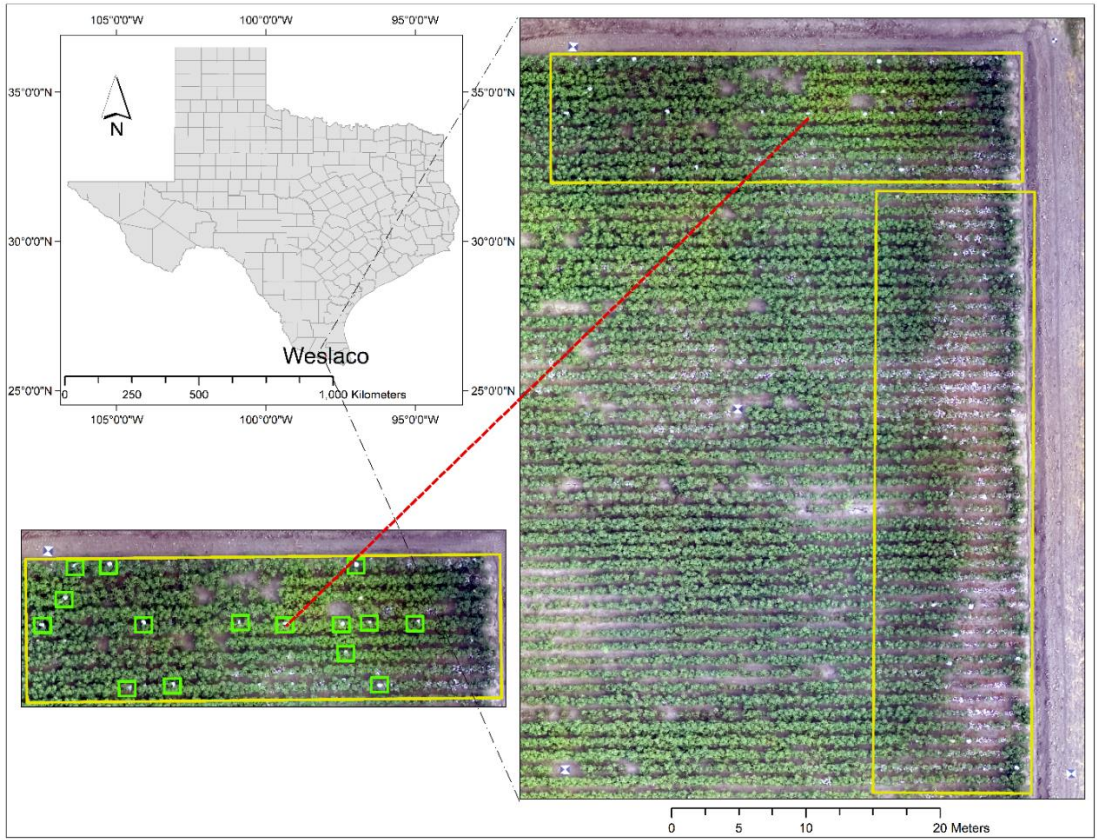

**Fig. 1.** An orthomosiac aerial image showing the experimental plot where white and brown color plastic shopping bags were tied on cotton plants as seen in yellow and green boxes.

## 2.2 Data Acquisition

A three band (RGB: red, green, and blue) FC6360 camera (Shenzhen DJI Sciences and Technologies Ltd., Shenzhen, Guangdong, China) integrated on a DJI Phantom P4 quadcopter (Shenzhen DJI Sciences and Technologies Ltd , Shenzhen, Guangdong, China) was used to collect aerial images of the test field from an altitude of 18.28 meter (60 feet) above ground level (AGL). Pix4DCapture (Pix4D S.A., Prilly, Switzerland) software was used on a regular smartphone running iOS version 13.5 (Apple Inc., Cupertino, California, U.S.A.) to control the quadcopter in flight. The camera has image resolution of 1600 x 1300 pixels. The 71 images collected had a ground sampling distance (GSD) of approximately 1 cm/pixel (0.40 inch/pixel). Data were collected on July 23, 2020, between 10 a.m. and 2 p.m. Central Standard Time (CST).

## 2.3 Image Data Preparation

After the aerial images were collected, they were used to generate an orthomosiac with Pix4DMapper 4.3.33 software (Pix4D S.A., Switzerland) (Fig. 1). The orthomosiac was used only as an aid in visually locating areas of interest on the ground. Individual RGB images of 1600 x 1300 pixels were split into 416 x 416 pixels with Python version 3.8.10 (Python Software Foundation, Delaware, United States) script (Yadav, 2021). Image augmentation techniques were subsequently applied to increase the number of images in the dataset with the Python library developed by Bloice et al. (2017). The following operations along with the corresponding probabilities were used in the augmentation pipeline: *rotate* with probability value of 0.7, *flip_left_right* with probability of 0.4, *zoom_random* with probability of 0.4, *percentage_area* of 0.8, and *flip_top_bottom* with probability of 0.4. These probability values determined the chance of applying the operation each time an image passed through the augmentation pipeline. Once an operation was chosen to be applied, parameter values were randomly applied from within the set range. 1000 samples were generated in each of the 5 iterations, resulting in a total of 5000 augmented images. However, many of these images did not contain either white or brown plastic bags. All such images were discarded, and only the images that contained at least one bag were used, resulting in a total of 1410 images. 10 sets of image data were allocated, with each set containing 141 images. Each of the 10 datasets was divided into training, validation and testing data in the ratio of 15:3:2. The LabelImg V-1.8.0 (Tzutalin, 2015) software tool was used for annotating ground truth bounding boxes for the two classes, white bags (*wb*) and brown bags (*bb*). All these processes can be seen in a workflow pipeline flowchart in Fig. 2.

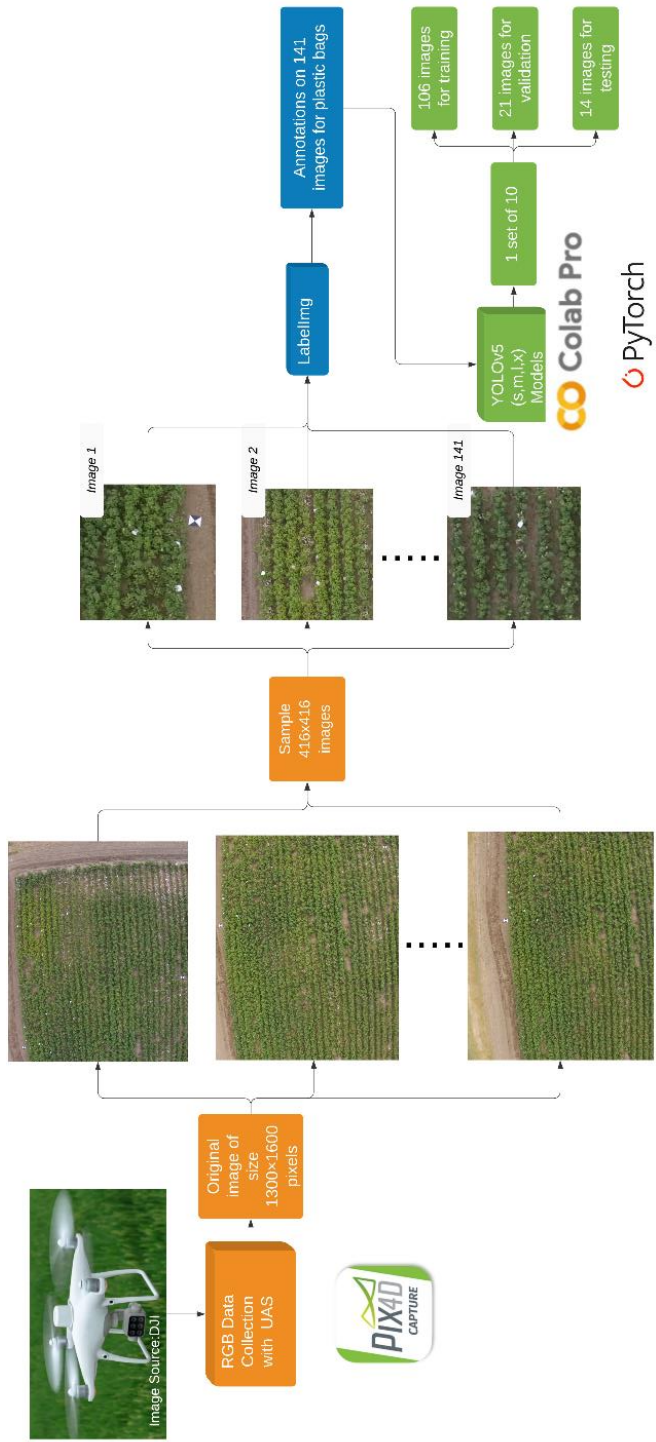

**Fig. 2.** Workflow pipeline showing the entire processes involved during the study from image data collection to YOLOv5 model training, validation, and testing.

**2.4 YOLOv5 Architecture**

All four versions of YOLOv5 have similar architectures that vary in depth and network parameters. To understand the function of YOLOv5, one can consider the architecture of YOLOv5s, shown in Fig. 3 as generated by the visualization software, Netron version 4.8.1 (Kist, 2021). The architecture has a total of 25 nodes, also known as modules (Yan et al., 2021), which are named *model/0* to *model/24*. The first 10 modules form the backbone network, the next 14 modules form the neck network, and the last module forms the head/detection network of the architecture. The backbone and neck network collectively form the feature extractor, while the head/detection network forms the predictor or detector.

The backbone network is comprised of focus, convolution, bottleneckCSP (Cross Stage Partial) and spatial pyramid pooling (SPP) modules. The focus module accepts input images of shape 3 x 640 x 640, where 3 represents the three channels (normally Red, Green, and Blue) while 640 x 640 represent image width and height in pixels. The focus module can process images of sizes other than 640 x 640 pixels and can be customized for different numbers of channels as well. The main objective of the focus module is to enhance the training speed as it makes use of the "hard-swish" activation function. This activation function is a modified version of the swish activation function, replacing the sigmoid function with a piecewise-linear "hard" equivalent (Equations 1 and 2) (Howard et al., 2019).

$$\text{swish(x)} = \text{x } \sigma(\text{x}) = \frac{\text{x}}{1 + \text{e}^{-\text{x}}} \tag{1}$$

$$h\text{-swish}(x) = x \frac{\text{ReLU6}(x+3)}{6} \tag{2}$$

where $\sigma$ is the sigmoid function and ReLU6 is a modified version of rectified linear unit (ReLU) activation function in which the maximum value is limited to 6. The bottleneckCSP module enhances the feature extraction process by making use of a residual network, i.e., concatenating deeper features with shallow features. The SPP module is used to ensure that a fixed-size feature vector is generated from any different size feature maps by using three parallel maxpooling layers.

The neck network is used to ensure that detection accuracy of objects is not compromised due to different scales and sizes. It makes use of a path aggregation network (PANet) to enhance feature propagation irrespective of scale (S. Liu et al., 2018). Modules 17, 20 and 23, which are part of the PANet and belong to the Neck network (also called P3, P4 and P5), output three feature maps belonging to objects at scales of small, medium, and large, respectively. P3 outputs a feature map of 80 x 80 pixels, while P4 and P5 output feature maps of 40 x 40 and 20 x 20, respectively.

The detect network is comprised of three detect layers to make detections at three scales corresponding to the features output from P3, P4 and P5. This network applies three anchor boxes at each scale on the three feature maps to output a vector containing information about the class probability, objectness score and predicted bounding box (BB) coordinates.

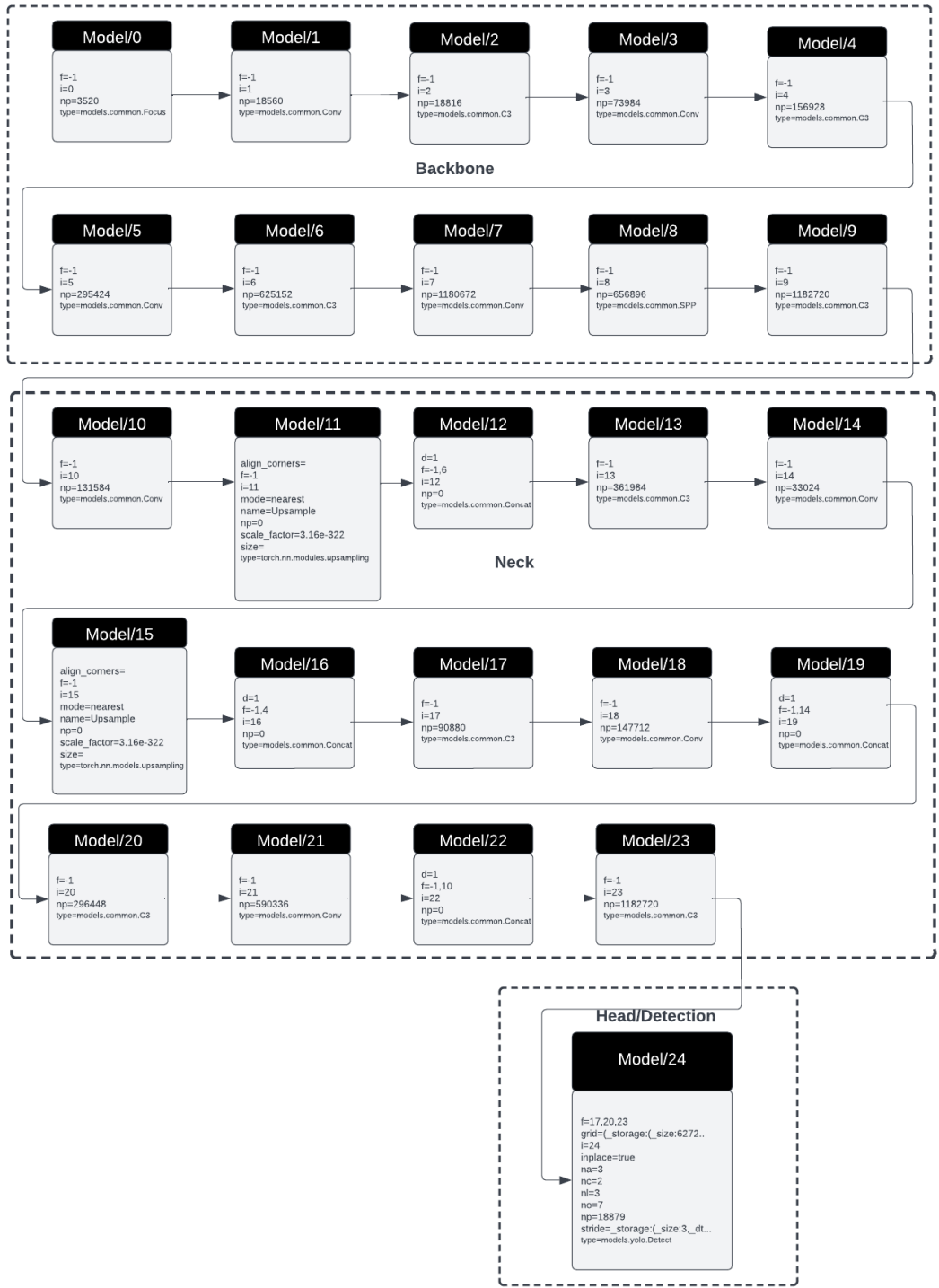

**Fig. 3.** Customized YOLOv5s network architecture for detecting 2 classes as generated by Netron visualization software.

## 2.5 YOLOv5 Training, Performance Metrics and Desirability Function

Source code for YOLOv5 was obtained from the Ultralytics Inc. GitHub repository of YOLOv5 (Jocher et al., 2021). Selected model hyperparameters are shown in Table 1. The PyTorch framework (Facebook AI Research Lab, Melno Park, CA, U.S.A.) with torch version 1.9.0 and Compute Unified Device Architecture (CUDA) version 10.2 (NVIDIA, Santa Clara, CA, U.S.A.) were used to implement YOLOv5. The auto anchoring function of YOLOv5 generated 4.5 anchors per target object based on the annotated ground truth bounding boxes (BBs). Original values from the source code were used for enhancement coefficients of hue (0.015), saturation (0.7) and lightness (0.4). Stochastic gradient descent (SGD) was used as the optimization algorithm for each version. SGD is a method of minimizing an objective function during model training by randomly choosing a single data point from a batch of data points, thereby reducing computation time to speed up the training process (Ruder, 2016). Google Colaboratory Pro (Google LLC, Mountain View, CA, U.S.A.) was used to train YOLOv5 on Tesla P100 GPU -16GB (NVIDIA, Santa Clara, CA, U.S.A.). All four versions of YOLOv5 were trained for 250 iterations with a batch size of 4, using 106 training, 21 validation, and 14 test images, resulting in a total step size of 6625. YOLOv5 was pretrained on COCO datasets for 80 different classes (Lin et al., 2014). The pretrained weights were used as the starting point, and YOLOv5 was customized for 2 classes based on a custom dataset. Each trained model was saved in *.pt* format and later used for detection.

**Table 1.**
YOLOv5 hyperparameters used for this study

| Hyperparameters | Values |
|---|---|
| Learning Rate | 0.01 |
| Learning rate Decay | 0.2 |
| Momentum | 0.937 |
| Weight Decay | 0.0005 |
| Batch Size | 4 |
| Training Epoch | 250 |

To evaluate the trained models on each of the ten datasets, precision, recall, average precision (AP), mean average precision (mAP@50), classification accuracy and F1-score were used as performance metrics. These metrics were calculated based on generalized intersection over union loss (GIoU loss), confidence over objectness loss, and classification over probability loss. The equations to calculate these metrics are given below.

$$P = \frac{TP}{TP+FP} \tag{3}$$

$$R = \frac{TP}{TP+FN} \tag{4}$$

$$Accuracy = \frac{TP+TN}{TP+TN+FP+FN} \tag{5}$$

$$F1\text{-}score = 2\frac{P*R}{P+R} \tag{6}$$

$$AP = \sum_{k=0}^{k=n-1}[R(k)-R(k+1)]\times P(k) \tag{7}$$

$$mAP = \frac{1}{n}\sum_{k=1}^{k=N} AP_k \tag{8}$$

where *P* is precision, *R* is recall, *N* is the number of classes (*N=2* in our case for *wb* and *bb*), *TP* is the number of true positives, *TN* is the number of true negatives, *FP* is the number of false positives, *FN* is the number of false negatives, $AP_k$ is the *AP* of class *k*, and *n* is the number of thresholds (*n=1* in our case). Precision is a measure of *TP* over predicted positives, while recall is a measure of *TP* rate and is also sometimes called sensitivity. Accuracy on the other hand is an overall measure of true detection. F1-score is the harmonic mean of *P* and *R* i.e., reciprocal of the arithmetic mean of reciprocals of *P* and *R*. All these metrics are determined based upon different losses that are calculated as shown below.

$$\text{GIoU}_{\text{loss}} = \lambda_{\text{coord}} \sum_{i=1}^{s^2} \sum_{j=1}^{B} l_{ij}{}^{\text{obj}} \; L_{\text{GIoU}} \tag{9}$$

Confidence Loss/ Objectness Loss=

$$\sum_{i=1}^{s^2} \sum_{j=1}^{B} l_{ij}^{\text{obj}} (C_i - \widehat{C}_i)^2 + \lambda_{\text{nobj}} \sum_{i=1}^{s^2} \sum_{j=1}^{B} l_{ij}^{\text{obj}} (C_i - \widehat{C}_i)^2 \tag{10}$$

Classification/Probability Loss=$\sum_{i=1}^{s^2} l_{ij}^{\text{obj}} \sum_{c \in \text{class}} (p_i(c) - \hat{p}_i(c))^2 \tag{11}$

In equation 9, $\lambda_{coord}$ is the penalty coefficient associated with the central coordinate of the predicted bounding box (BB), $s^2$ is the number of grid cells generated on the input image, and B is the total number of predefined BB, i.e., the anchors which is equal to 9 (three at each scale). In equations 10 and 11, existence and absence of an object in the predicted BB are represented by '*obj*' and '*nobj*', respectively. Similarly, $C_i$ represents class of the predicted object, $\widehat{C}_i$ represents class of the ground truth object, and $\lambda_{\text{nobj}}$ represents the penalty coefficient associated with the confidence loss. Even though these metrics are widely used for evaluating performance of the trained model and its associated classifier, one must be careful about two important aspects: the way data are sampled and the balance between different classes in sampled datasets. Xie and Braga-Neto (2019) found that precision as a performance metric may not be reliable and can be severely biased under separate sampling conditions. The sampling technique used in this research was kept consistent to avoid bias induced by sampling discrepancy. Similarly, Luque et al. ( 2019) showed that imbalance in the proportion of instances between positive and negative classes can affect classification accuracy, and they proposed geometric mean (GM) or Bookmaker Informedness as the most bias-free metrics. To solve this issue YOLOv5 also uses focal loss (Li et al., 2018).

In addition to evaluating each model's performance separately, the best model had to be identified among all four variants (*s, m, l,* and *x*) based on the performance metrics and inference time. Identifying the best overall model would enable its deployment on a UAS using CV for real-time plastic bag detection in a cotton field. Since this problem is similar to optimization with multiple response variables, the desirability function was used to convert all the response variables into a single one by first scaling them from 0 to 1, then taking their geometric mean (GM), and finally combining them as shown in equation 12 (Costa et al., 2011; Obermiller, 2000).

$$D = \sqrt[p]{(d_1 \text{ x } d_2 \text{ x } d_3 \text{ x} \ldots d_p)} \tag{12}$$

Here, *D* is the overall desirability of all the *p* response variables (in our case *p* = 4) denoted by $d_1, d_2 \ldots d_p$. In our case, $d_1 = $ map@50, $d_2 = $ accuracy of white bags, $d_3 = $ accuracy of brown bags and $d_4 = $ inference speed. To implement this function, JMP Pro version 15.2.0 software was used (SAS Institute, North Carolina, U.S.A.). In this study, the response goals for mAP@50 and accuracies for white and brown bags were set to 1, 0.85 and 0.8 for high, middle, and low respectively and then the desirability was maximized using constrained Newton's method as explained by Dennis and Schnabel (1996).

## 2.6 Experiment Design

To achieve the second objective, a fixed effect model was designed to examine the effect of the color of plastic shopping bags on AP and accuracy, with all four variants of YOLOv5 model. Ten observations were made by randomly selecting ten sets of 14 test images containing both white and brown plastic shopping bags and then attempting to detect them with the trained YOLOv5s, YOLOv5m, YOLOv5l and YOLOv5x models. Assuming $Y_{ijc}$ is the *j-th* observation from treatment group *ic*, where $ic$ = W, B, and $jc$ = 1,2,3, …10, the fixed effect model can be expressed as:

$$Y_{ijc} = \mu_{.c} + \alpha_{ic} + \varepsilon_{ijc} \tag{13}$$

Here, $\mu_{.c}$ is the average of all the treatment group means, while $\varepsilon_{ijc}$ is the error, or the residual term, for the fixed effect model and $\alpha_{ic}$ is the *i-th* treatment effect. The null and alternative hypotheses for this model can be shown as:

$$H_{0c}: \alpha_W = \alpha_B \tag{14}$$

$$H_{1c}: \alpha_W \neq \alpha_B \tag{15}$$

The first step was to test the normality assumption with the Shapiro-Wilk test at the 95% confidence level ($\alpha = 0.05$) with Python's *SciPy Stats* module (The SciPy Community, 2021). Then the model effect test was conducted with the Standard Least Squares (SLS) personality fit with JMP Pro version 15.2.0 software (SAS Institute, North

Carolina, U.S.A.). The SLS method is used to fit linear models for continuous-response variables by using the method of least squares.

To achieve the third objective, a fixed effect model was designed to test the effects of YOLOv5 versions on the AP, mAP@50, and inference speed for white and brown bags. Again, there were 10 observations, 4 treatment groups (YOLOv5s, YOLOv5m, YOLOv5l and YOLOv5x) and 4 output metrics (AP for white bags, AP for brown bags, mAP@50, and inference speed). Assuming $Y_{ij}$ is the *j-th* observation from treatment group *I*, where *i* = s, m, l, x, and *j*= 1,2,3, …10, the fixed effect model can be expressed as:

$$Y_{ij} = \mu_{.} + \alpha_i + \varepsilon_{ij} \tag{16}$$

Here, $\mu_{.}$ is the average of all the treatment group means, while $\varepsilon_{ij}$ is the error (the residual term) for the fixed effect model, and $\alpha_i$ is the *i-th* treatment effect. The null and alternate hypotheses for this model can be shown as:

$$H_0: \alpha_s = \alpha_m = \alpha_l = \alpha_x \tag{17}$$

$$H_1: \alpha_s \neq \alpha_m \text{ or } \alpha_s \neq \alpha_l \text{ or } \alpha_s \neq \alpha_l \text{ or } \alpha_m \neq \alpha_l \text{ or } \alpha_m \neq \alpha_x \text{ or } \alpha_l \neq \alpha_x \tag{18}$$

Like the previous case, the normality assumption was first tested and then the model effect, with SLS personality fit with JMP Pro version 15.2.0 software (SAS Institute, North Carolina, U.S.A.). Finally, to choose the optimal YOLOv5 model, the desirability function in the prediction profiler graphs obtained from the YOLOv5 model type effect tests was used.

Once the most desirable variant of YOLOv5 was chosen, a fixed effect model test was performed to determine the effect of height of plastic bags (top, middle and bottom) on the overall detection accuracy by the trained YOLOv5 model. For this 10 random image samples (416 x 416 pixels) were chosen and the most desirable variant of the trained YOLOv5 model was used for detection of plastic bags. The amount of detected bags was matched visually with the experiment layout to determine the amount of top, middle and bottom bags detected out of the total top, middle and bottom bags present in each of the image. Then a percentage ratio for each (top, middle and bottom) was calculated. This is called bag-based accuracy hereon.

**3.0 Results**

**3.1 Overall Performance of YOLOv5 for Plastic Bag Detection in Cotton Field**

Fig. 4. shows the three types of losses that were calculated and plotted for training and validation datasets.

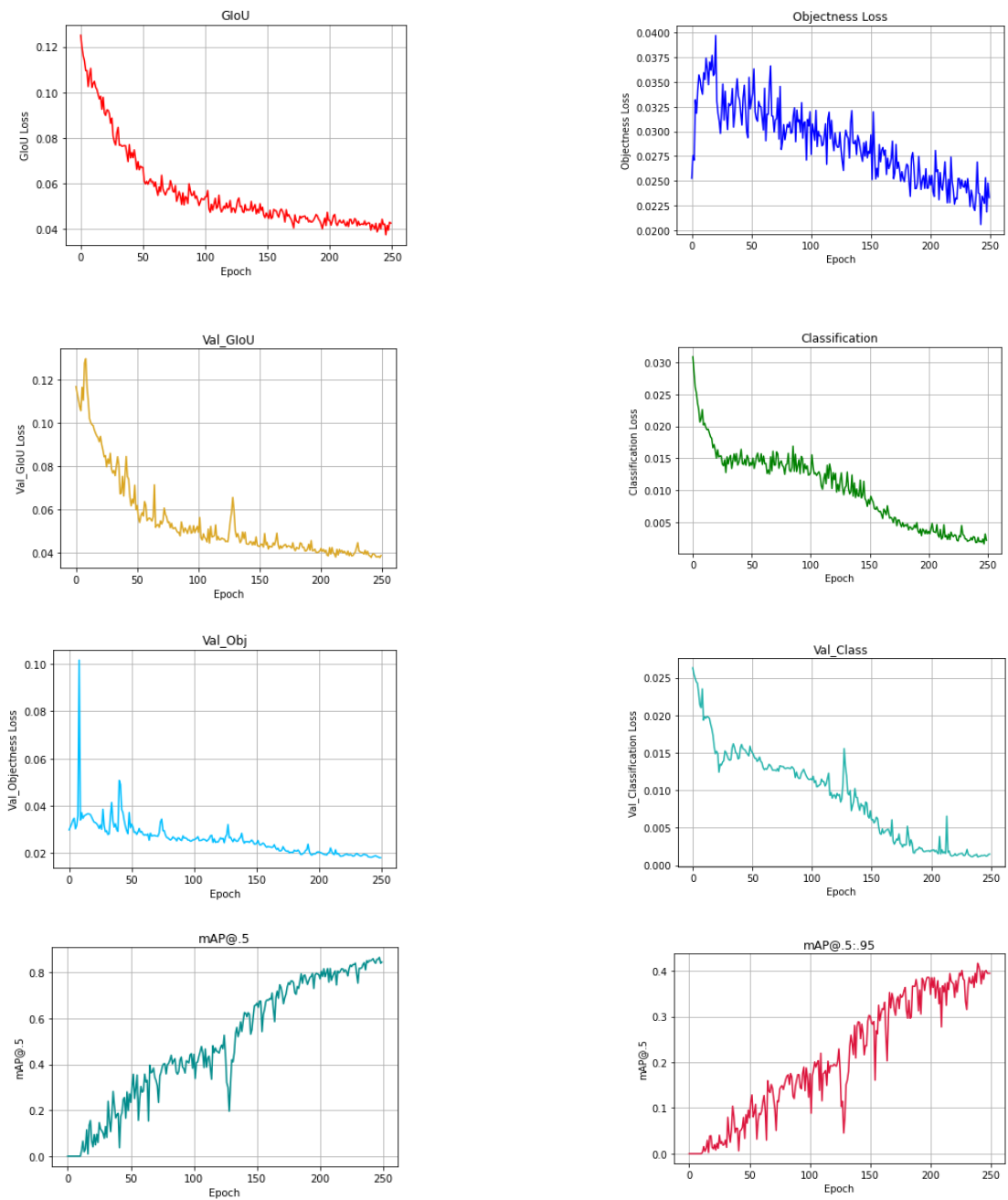

**Fig. 4.** Example plots showing different losses and performance metrics based on training and validation datasets for YOLOv5s.

The plot titles with *val_* as the prefix were obtained from the validation dataset. Based on the 40 total observations (10 observations from four variants of YOLOv5), the mean AP for white bags was 90.53% with a standard deviation of 0.08. The mean AP for brown bags was 84.84% with a standard deviation of 0.06. The mean of mAP@50 was found to be 87.68% with a standard deviation of 0.05. The accuracy for white bags had a mean of 92.35% and standard deviation of 0.05. The mean accuracy for brown bags was 77.87% with a standard deviation of 0.11. The mean inference speed was 81.43 FPS with a standard deviation of 39.78. Fig. 5(A) is a precision-recall plot obtained after training YOLOv5s with one of the 10 observation datasets. The areas under these curves can be used to determine AP for white bags (*wb*) and brown bags (*bb*) as well as mAP@50 corresponding to both. Fig. 5(B) includes plots of F1-score against confidence values, which are directly proportional up to a confidence value of around 0.33. After that point, the F1-score starts to decrease with increasing confidence values, indicating that the predicted BBs can be associated with more confidence about the presence of included objects, but also that they are less precise and accurate for predicted classes (*wb* and *bb*). Until the confidence value of approximately 0.62, the predicted BBs are more accurate, precise, and sensitive about *wb*, but after this point they become more precise, sensitive, and accurate about *bb*. Fig. 6 is a confusion matrix heat map in which white bags have a higher true positive rate than the brown bags. Fig. 7 includes some results of the trained model used for detection on the test dataset.

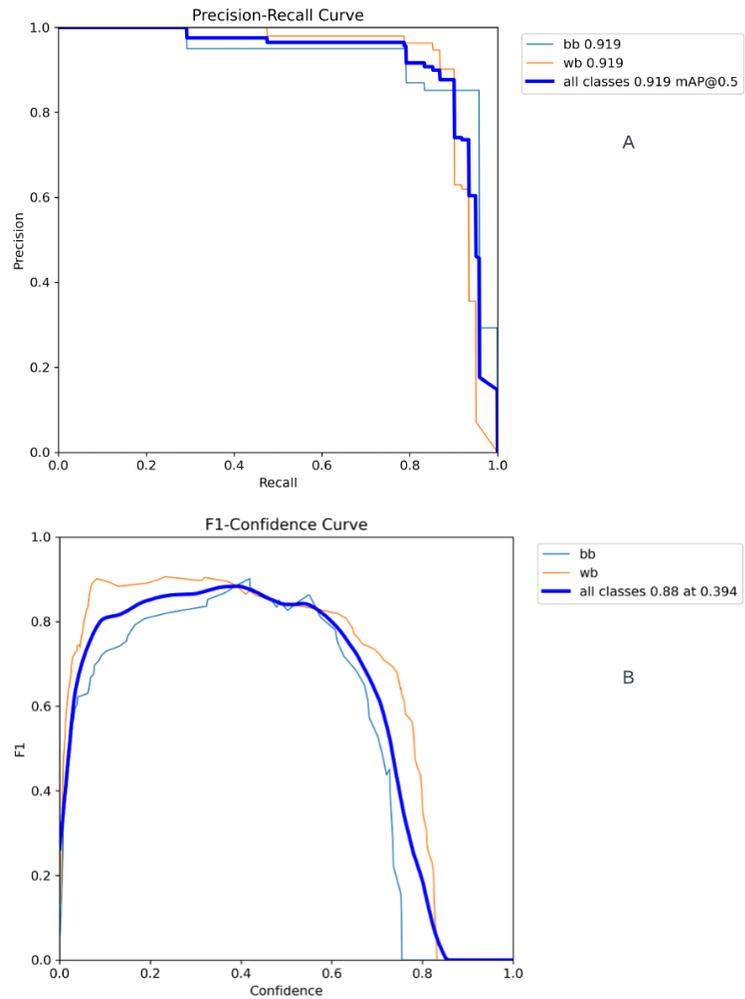

**Fig. 5.** (A) Precision Recall curve that was obtained after one of the training processes. (B) Example of F1-score values plotted against confidence scores.

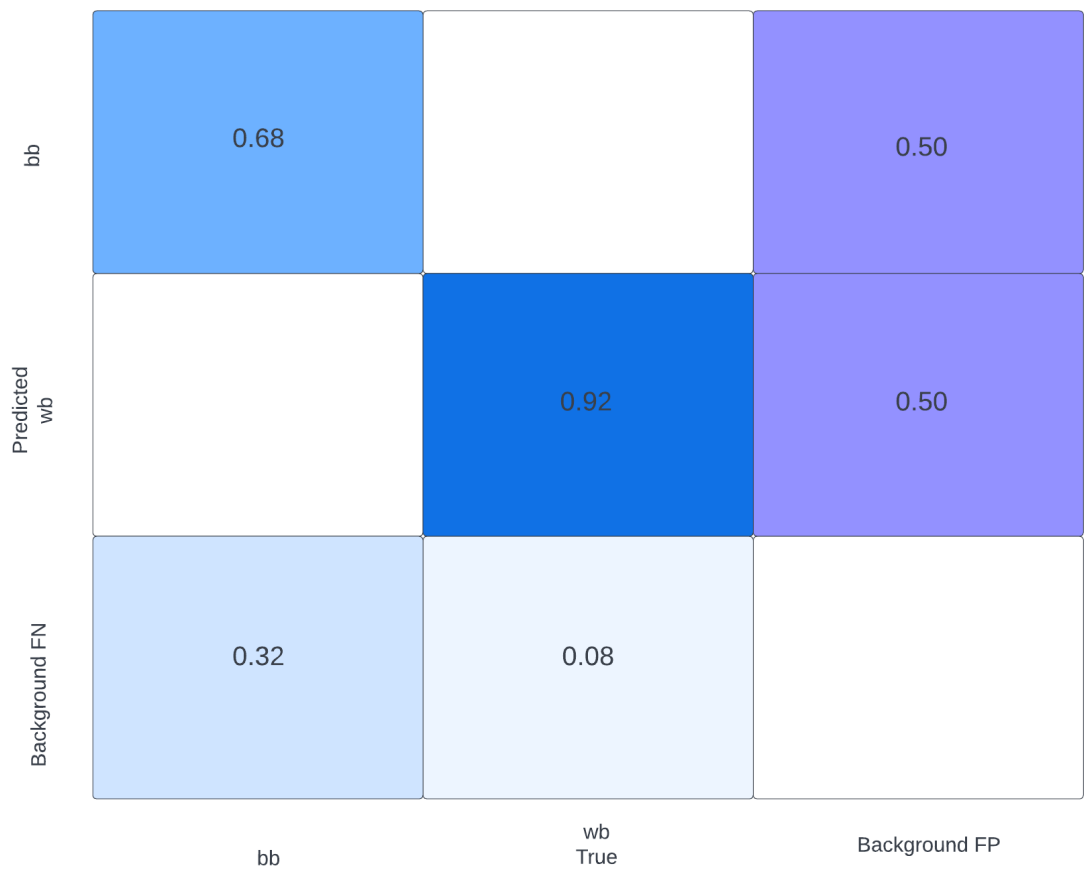

**Fig. 6.** Example of confusion matrix obtained after training YOLOv5s and using one of the 10 validation datasets.

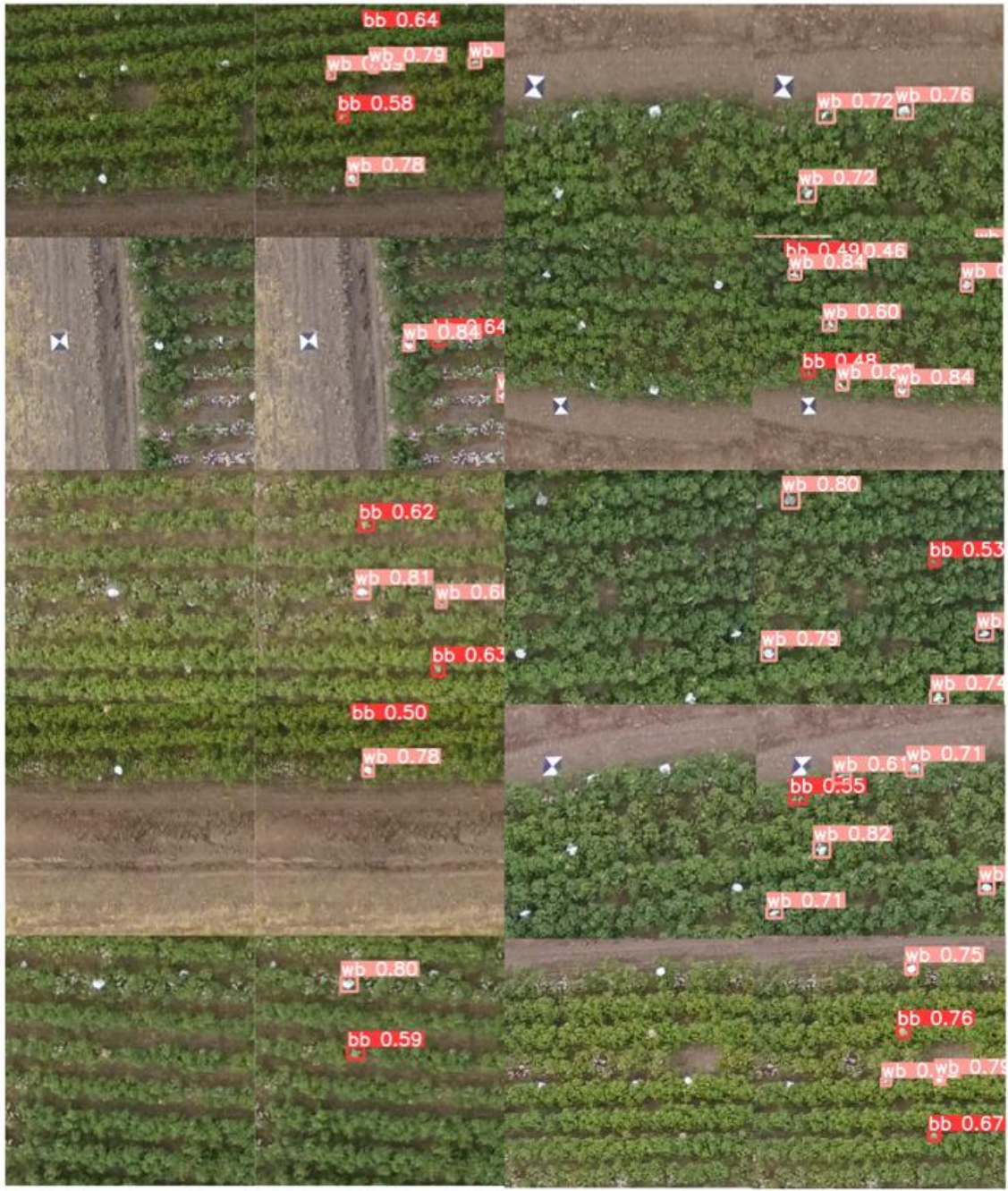

**Fig. 7.** Example images showing white and brown plastic bags being detected by trained YOLOv5s within predicted BBs and the corresponding class confidence scores.

## 3.2 Color Effect Tests

When the color effect model was tested on the four versions of YOLOv5, the resulting histogram and density plots (Figs. 8, 9, 10 and 11) showed the distribution of AP and accuracy for each version (*s, m, l,* and *x*) as well as side-by-side box plots with lines connecting their means and medians along with pair-wise Student's *t*-test values. Based on 10 observations for both white and brown bags, the mean of AP values for *s, m, l,* and *x* models were 89.85%, 90.15%, 87.94% and 82.81% with standard deviations of 0.06, 0.05, 0.07 and 0.11 respectively. The overall mean accuracies for *s, m, l,* and *x* were 84.80%, 88.95%, 85.85% and 80.85% respectively. Based on colors, both the AP and accuracy for white bags were significantly higher than the brown bags for *s* and *l* (Figs. 8 and 10). However, for *m* and *x*, only the accuracy for white bags was significantly higher than the brown bags (Figs. 9 and 11). The AP and accuracies were expected to be higher with increasing size and depth of the model, but this was true only from *s* to *m*, not from *m* to *l* or *l* to *x*. A possible reason for this could be that there were fixed numbers of training iterations for all versions. Hence, the *l* and *x* versions may not have reached convergence during the 250 iterations. In this study it was necessary to fix the hyperparameter values for consistency in results comparisons. Table 2 shows the results of the effect of color of plastic bags on AP and accuracy for all four model versions. There was evidence to determine that color had a significant effect ($\alpha = 0.05$) on accuracy for all four versions. Similarly, there was evidence to determine that color had a significant effect on AP for *s* and *l* but not for *m* and *x* at the 5% significance level. From the boxplots of accuracy as a function of color of bags, white bags had significantly higher accuracy than brown bags (Fig. 12).

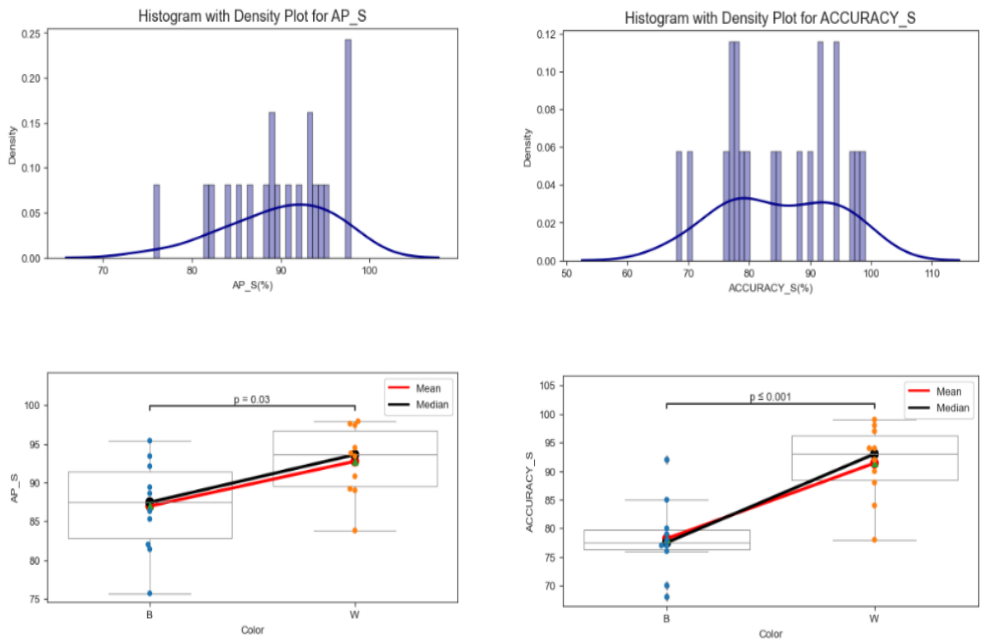

**Fig. 8.** Distribution of average precision (AP) and accuracy for YOLOv5s as well as for white and brown color plastic shopping bags.

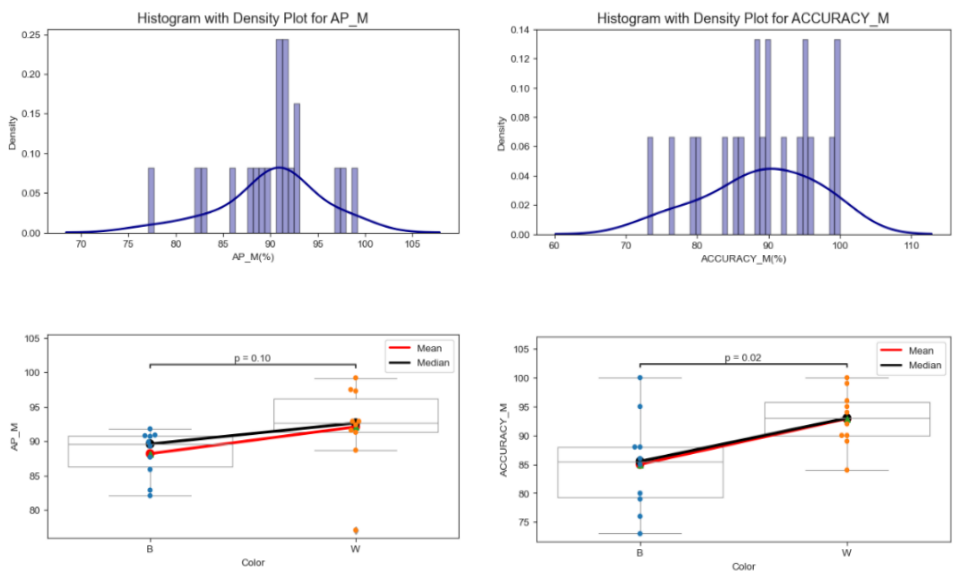

**Fig. 9.** Distribution of average precision (AP) and accuracy for YOLOv5m as well as for white and brown color plastic shopping bags.

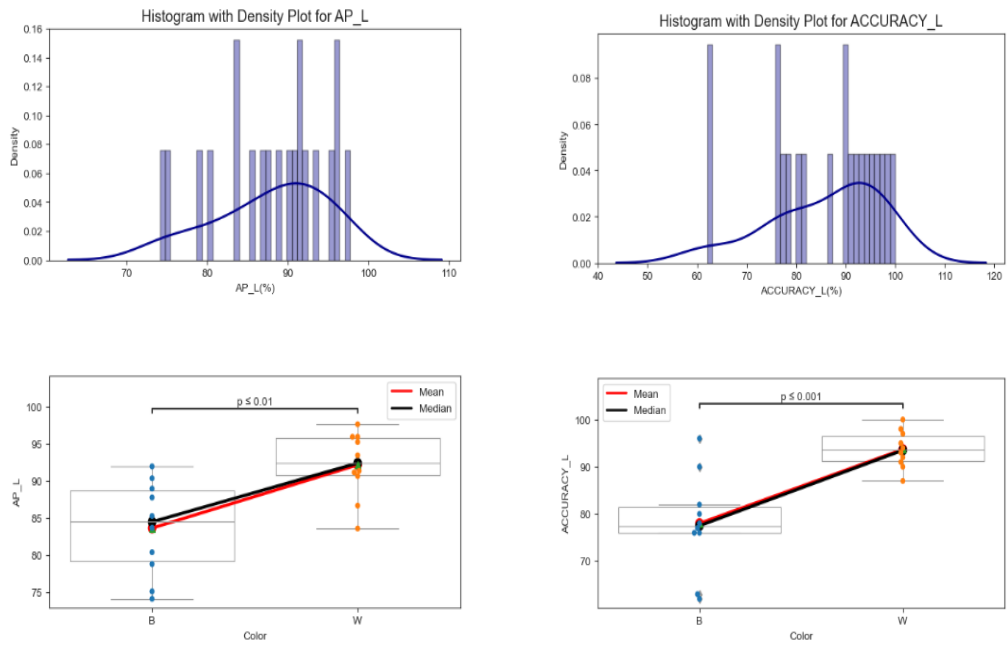

**Fig. 10.** Distribution of average precision (AP) and accuracy for YOLOv5l as well as for white and brown color plastic shopping bags.

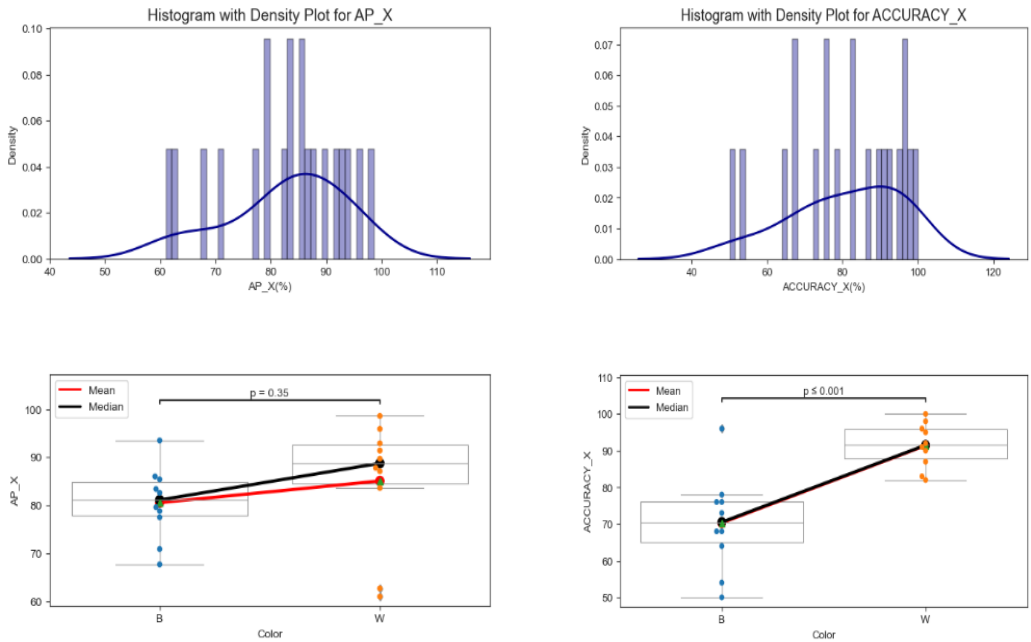

**Fig. 11.** Distribution of average precision (AP) and accuracy for YOLOv5x as well as for white and brown color plastic shopping bags.

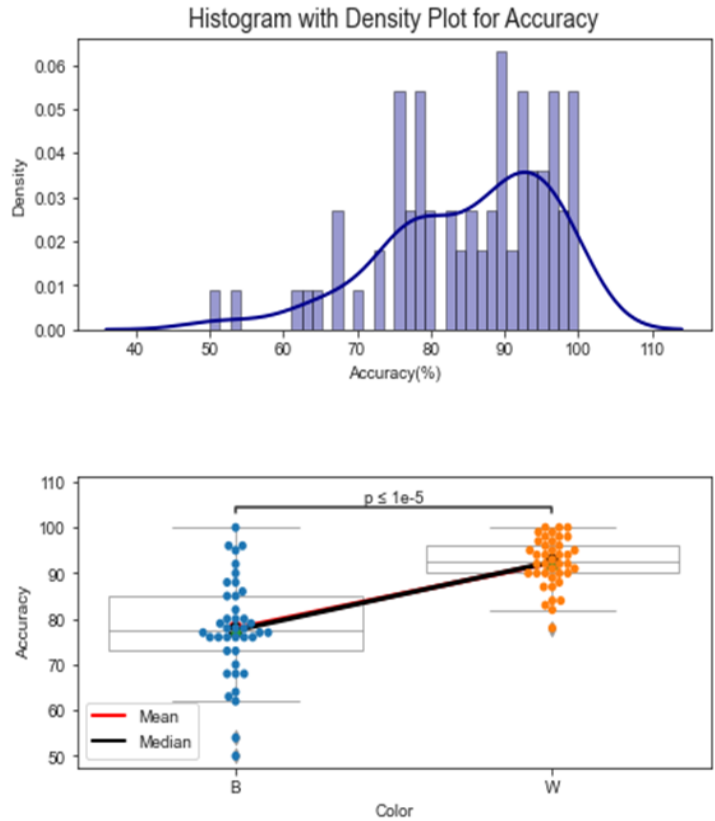

**Fig. 12.** Distribution of overall accuracy of the YOLOv5 models as well as accuracy for white and brown color plastic shopping bags.

**Table 2.** Plastic bag color effect test on average precision and accuracy of different variants of YOLOv5.

| Model | Average Precision | | Accuracy | |
|---|---|---|---|---|
| | F Ratio | Prob>F | F Ratio | Prob>F |
| YOLOv5s | 5.7812 | 0.0272 | 19.2649 | 0.0004 |
| YOLOv5m | 3.0022 | 0.1002 | 6.6480 | 0.0189 |
| YOLOv5l | 12.1663 | 0.0026 | 19.8409 | 0.0003 |
| YOLOv5x | 0.9136 | 0.3518 | 21.6296 | 0.0002 |

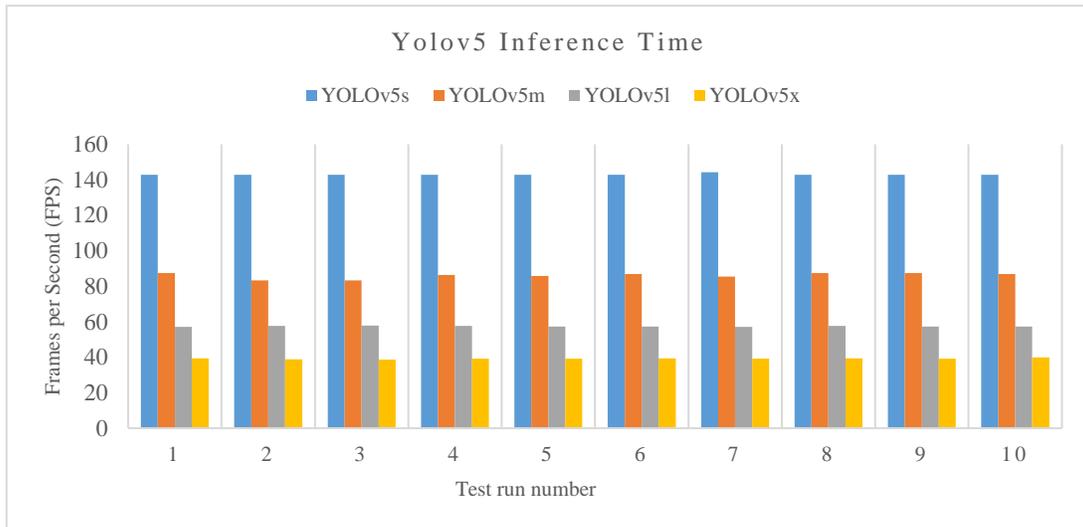

**Fig. 13.** A chart showing difference inference times that were obtained by testing 14 images during each of the 10-test dataset.

### 3.3 Model type effect tests

Values of AP, accuracies, mAP@50, and inference times for the model effect test are given in Table 3. Model type did not have a significant effect on accuracy and AP for white bags, but it did have a significant effect in all the other metrics ($\alpha = 0.05$). In part this means that detection of white bags was not affected by the choice of model, whereas detection of brown bags was affected by the type of model used. Furthermore, inference time (i.e., detection speed) depends on choice of model regardless of bag color. This result was expected because larger models are deeper with many more parameters that require more processing time, as evidenced from the different inference times obtained for each observation test dataset and for all four versions (Fig. 13). The distribution of mAP@50 for the four YOLOv5 version models can be seen in Fig. 14, with histogram and density plots as well as side-by-side box plots with means and medians connected by lines. Pair-wise Student's *t*-test values are also shown with the boxplots. The most consistent distribution can be seen for YOLOv5m, but there was an outlier. The mean and median lines are almost parallel between YOLOv5s and YOLOv5m, indicating similar performance. The lines decline from YOLOv5m to YOLOv5l and from YOLOv5l to YOLOv5x, indicating a sequential decrease in the mAP@50 for the larger models. As mentioned earlier, this result can be attributed to the fact that the larger models were likely not trained adequately for their losses to converge. Mean accuracy increased from *s* to *l* but decreased from *l* to *x*; however, the decrease is not significant. Model type did not have a significant effect on overall accuracy except between *m* and *x* (Fig. 14).

**Table 3.**
YOLOv5 Model type effect test on AP, accuracy, and mAP@50 for white and brown color plastic bags.

| Model | | | White Bags | | mAP@50 | | Brown Bags | |
|---|---|---|---|---|---|---|---|---|
| | F Ratio | Prob>F | F Ratio | Prob>F | F Ratio | Prob>F | F Ratio | Prob>F |
| AP | | | 2.15 | 0.11 | 4.63 | 0.01 | 3.27 | 0.03 |
| Accuracy | | | 0.44 | 0.73 | | | 3.66 | 0.02 |
| Inference Time | 27665.3 | <0.0001 | | | | | | |

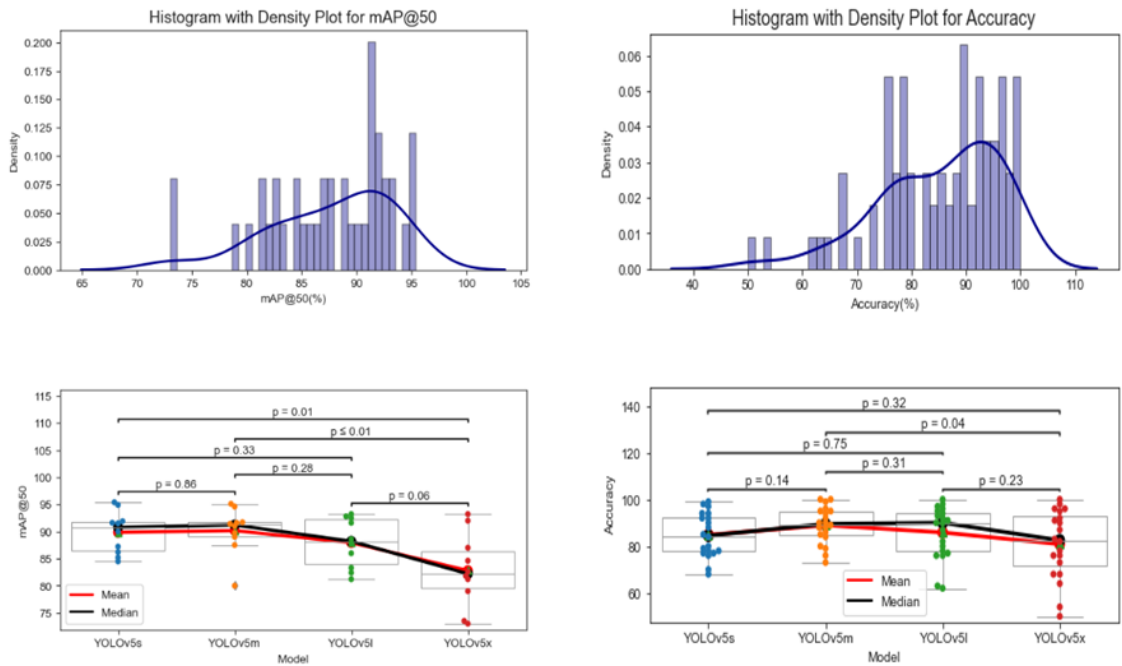

**Fig. 14.** Distribution of accuracies and mean average precision (mAP@50) for all the four variants of YOLOv5 model.

### 3.5 Desirability function as optimization method

The most important of the six metrics (AP and accuracy for white and brown bags, mAP@50, and inference time) were the inference speed, accuracy, and mAP@50, because the goal was to find an optimal YOLOv5 model with greatest detection accuracies for white and brown plastic bags at a higher inference speed that can be deployed on a GPU for near real-time detection. The maximum desirability obtained was nearly 0.95 for YOLOv5m, which had the maximum inference speed of 86 FPS, accuracies for white and brown bags greater than 92% and 85% respectively and mAP@50 greater than 90% (Fig. 15).

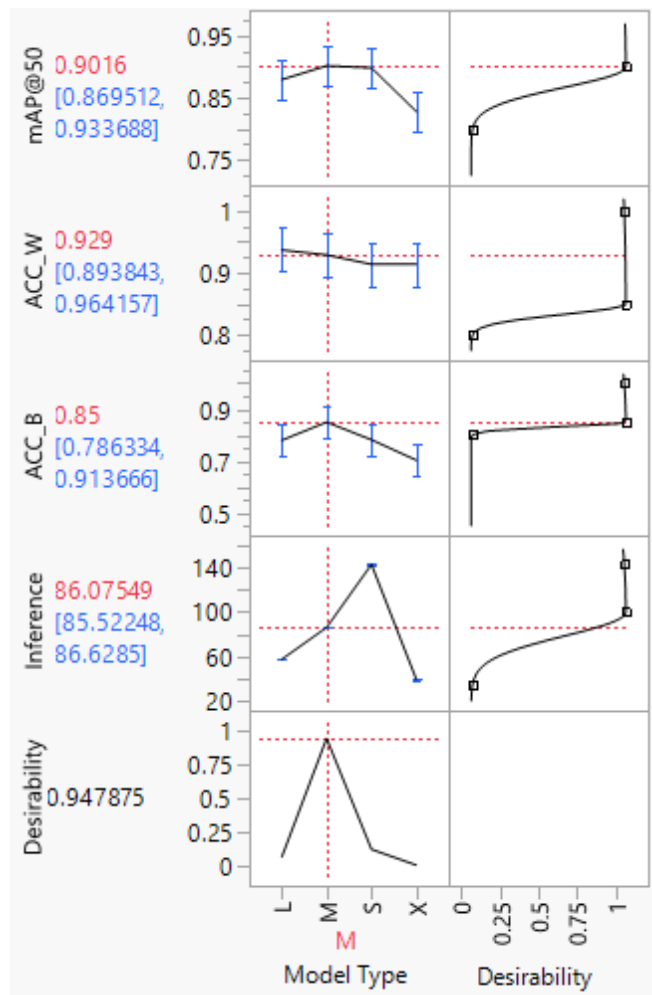

**Fig. 14.** Prediction profilers showing maximum desirability for the YOLOv5m (M) model.

**3.4 Plastic bag height effect test**

Once the most desirable variant of YOLOv5 was found i.e., YOLOv5m, it was used to test how the height of plastic bags on cotton plants affects its detection accuracy (bag-based detection accuracy). Bag-based detection accuracies i.e., the percentage of top, middle and bottom bags that were detected in each of the ten test images are shown in Table 4.

**Table 4**

Effect test of plastic bag height on bag-based detection accuracy. Top, Middle and Bottom show the percentage of corresponding bags detected in each of the ten test images on which the trained YOLOv5m model was used.

| Image_Sample_Number | Top | Middle | Bottom |
|---|---|---|---|
| 1 | 62.5 | 12.5 | 0 |
| 2 | 100 | 0 | 0 |
| 3 | 100 | 100 | 0 |
| 4 | 100 | 0 | 0 |
| 5 | 100 | 0 | 0 |
| 6 | 100 | 33.33 | 0 |
| 7 | 100 | 100 | 50 |
| 8 | 100 | 50 | 0 |
| 9 | 100 | 100 | 0 |
| 10 | 80 | 100 | 0 |

It was found that on an average 94.25% of the top bags (white and brown), 49.58% of the middle bags (white and brown) and 5% of the bottom bags (white and brown) were detected by the trained YOLOv5m model. A fixed effect model test determined that height of plastic bags had a significant effect ($p < 0.0001$) on bag-based detection accuracy.

**4.0 Discussion**

**4.1 Overall performance of YOLOv5**

Results from this study showed that YOLOv5 was able to detect plastic shopping bags in aerial images of a cotton field with an overall accuracy of 92% for white bags, 78% for brown bags and mAP@50 of 88%. This means using the deep learning method, detection accuracy for white bags improved by nearly 19%, for brown bags by 18% and the overall detection accuracy improved by 24% when compared to the previous methods used (Yadav et al., 2020). Apart from higher detection accuracy, this method was also able to perform detection at speeds that make it practically viable for near real-time detection (Fig. 13). Therefore, this approach can be used to potentially enable field robots like unmanned ground vehicles to remove the bags autonomously as proposed by Hardin et al. (Hardin et al., 2018). It was also found that the overall detection accuracy of white bags was significantly higher than that of the brown bags. This result was expected because the brown bags had lower contrast with their surroundings, particularly those that were tied at the bottom of the plants and looked like the soil in the background. This is in agreement to the findings by Liu et al. (2021) in which detection of aircraft by YOLOv5 decreased due to white background as most of the aircrafts are white in color.

It is known that the deeper the CNN network, the better the performance. Deeper networks learn better (Bevilacqua et al., 2019; Raghavendra et al., 2018), explaining why in most cases the performance of YOLOv5x > YOLOv5l > YOLOv5m > YOLOv5s (Solawetz, Jacob;Nelson, 2020). Results of our study showed this to be true for YOLOv5s and YOLOv5m, but the performance was found to be less for YOLOv5l and YOLOv5x. There can be multiple reasons for this, one of which may be attributed to the fact   the larger and deeper models were not trained enough to enable the losses to converge within the fixed 250 iterations. DL and ML models are stochastic in nature; hence fine tuning hyperparameters values of the $l$ and $x$ other than the ones in Table 1 can result in different and perhaps higher detection results. However, for fair comparison, all the four versions were trained with the same number of iterations and fixed values of hyperparameters as explained in sections 5 and 6.

## 4.2 Effect of color of bags on YOLOv5 performance

From this study, it was found that color of plastic bags has a significant effect on overall detection accuracy for all four variants of YOLOv5. White bags were always detected with higher accuracy than brown bags, as was expected due to its higher contrast with the background. This result could be different if the cotton plants had many open bolls, thus creating a white background and making detection of white bags difficult due to its decreased contrast with the background. An interesting finding was that bag color had a significant effect on AP for YOLOv5s and YOLOv5l but not for the YOLOv5m and YOLOv5x. This means that the $m$ and $x$ models were fine-tuned and trained such that the AP of these models were robust to the changes in color of plastic bags.

### 4.3 Effect of model type

Findings from this study indicated that YOLOv5m performed the best in terms of overall accuracy and mAP followed by YOLOv5s, YOLOv5l and YOLOv5x. However, no significant effect was found on the overall detection accuracy of plastic bags except between YOLOv5m and YOLOv5x. These results suggest that if any of the models is trained adequately, it can be used for plastic bag detection in cotton fields without major concerns about detection accuracy. This is only true if computation cost is not an issue which in practical situation is always a matter of concern.

### 4.4 Optimal model

Computation cost in terms of time, energy to operate the system, money, etc. has always been a concern when deploying CV algorithms on edge computing devices (Dandois & Ellis, 2013; Samie et al., 2016). In our study, we also focused on finding the optimal YOLOv5 model out of the four variants keeping the detection accuracies for white and brown bags at least 80% and inference speed of at least 35 FPS. The desirability function as explained in section 2.5 was helpful to determine the most practically viable model for near real-time detection of plastic shopping bags in cottonfields which is YOLOv5m.

**4.5 Effect of plastic bag height**

The bag-based detection accuracies were found to be relatively consistent for the top and bottom bags with standard deviation of 12.80 and 15.81 respectively; however, in the case of middle bags, it was highly inconsistent with a standard deviation of 46.17. On an average, the trained YOLOv5m was able to detect top bags at significantly higher detection accuracy than the bottom bags. This is as expected because most of the bags at the bottom remined hidden under the canopied of cotton plants and therefore couldn't be seen in the aerial images collected by the UAS. The middle bags were detected at higher accuracy than the bottom bags but at a significantly lower value than the top bags.

**4.6 Limitations**

The findings of this study are limited by the fact that fixed hyperparameters and training iterations were used for all YOLOv5 variants, such that deeper models like YOLOv5l and YOLOv5x apparently failed to converge their losses. These models would likely perform better if trained for a greater number of iterations. Furthermore, the inference speed reported here is for a GPU that is part of high-performance computing system that is not readily available for practical applications in field conditions. Hence, the inference speed may be significantly different for a field-based computing platform.

**5.0 Conclusions**

In this paper, we were able to show that YOLOv5 can be used to detect white and brown color plastic shopping bags in a cottonfield on RGB aerial images collected by UAS. We were able to show that YOLOv5 detected the bags at greater accuracies than the classical ML methods used previously. We were also able to show that color of bags has significant effect on detection accuracies and YOLOv5m was found to be the most practically viable model for near real-time detection. In the end we were able to show that, overall, the top bags were detected at a significantly higher detection accuracy than the middle and bottom bags.

## 6.0 Future directions

Since YOLOv5 is built upon a PyTorch (Facebook AI Research Lab, Melno Park, CA, U.S.A.) framework, which is written in Python programming language, it is easy to deploy on embedded systems like GPUs. The inference results (Fig. 9) are promising for real-time object detection. However, these were obtained on Tesla P100 GPU -16GB (NVIDIA, Santa Clara, CA, U.S.A.), which provides computation speed not possible to deploy at "the edge" (i.e., at the source of image data generation, a UAS with camera and GPU). The goal of this research is to deploy the developed AI model on a low-cost and lightweight edge computing device that can be easily mounted on a UAS. Developing near real-time computation on an edge computing device will be part of future research. The future work will also involve extraction of central coordinates of detected BBs with plastic shopping bags and then georeferencing them to a geodetic coordinate system that might allow ground-based robots to collect the detected plastic shopping bags. Ideally, detection and coordinate extraction will be performed in near real time with the trained YOLOV5m model on a low cost, light weight GPU device mounted along with a camera on a UAS.

## Acknowledgement

This project was supported by Cotton Incorporated Inc. and Texas A& M AgriLife Research. We would like to extend our sincere thanks to  all the personnel involved including Roy Graves, Madison Hodges, Dr. Thiago Marconi, Jorge Solórzano Diaz, and Uriel Cholula for their contribution during field work . We also extend our sincere gratitude and thanks towards all the reviewers and editors.

*CRediT authorship contribution statement*

**Pappu Kumar Yadav**: Conceptualization, Data curation, Formal analysis, Investigation, Methodology, Validation, Visualization, Writing – original draft, Writing - review & editing. **J. Alex Thomasson:** Conceptualization, Data curation, Investigation, Methodology, Formal analysis, Project administration, Resources, Supervision, Writing - review & editing. **Robert Hardin:** Project administration, Formal analysis, Resources, Supervision, Writing - review & editing. **Stephen W. Searcy:** Formal analysis ,Writing - review & editing. **Ulisses Braga-Neto:** Investigation, Methodology, Formal analysis ,Writing - review & editing. **Sorin Popescu:** Formal analysis ,Writing - review & editing. **Roberto Rodriguez:** Data curation ,Formal analysis ,Writing - review & editing. **Daniel E Martin:** Writing - review & editing. **Juan Enciso:** Resources, Writing - review & editing. **Karem Meza:** Resources, Writing - review & editing. **Emma L. White:** Resources, Writing - review & editing.

**Declaration of Competing Interest**

The authors declare that they have no known competing financial interests or personal relationships that could have appeared to influence the work reported in this paper.

**References**

Bevilacqua, V., Brunetti, A., Guerriero, A., Trotta, G. F., Telegrafo, M., & Moschetta, M. (2019). A performance comparison between shallow and deeper neural networks supervised classification of tomosynthesis breast lesions images. *Cognitive Systems Research*, *53*, 3–19. https://doi.org/10.1016/j.cogsys.2018.04.011

Blake, C; Sui, R; Yang, C. (2020). UAV-Based Multispectral Detection of Plastic Debris In Cotton Fields. *2020 Beltwide Cotton Conferences, Austin, TX*, *53*(9), 560–564.

Brungel, R., & Friedrich, C. M. (2021). DETR and YOLOv5 : Exploring Performance and Self-Training for Diabetic Foot Ulcer Detection. *34th International Symposium on Computer-Based Medical Systems (CBMS) 2021*, 148–153. https://doi.org/10.1109/CBMS52027.2021.00063

Chen, Y. Q. (1995). Novel Techniques for Image Texture Classification. In *Thesis* (Issue March). University of Southampton.

Chen, Y. R., Chao, K., & Kim, M. S. (2002). Machine vision technology for agricultural applications. *Computers and Electronics in Agriculture*, *36*(2–3), 173–191. https://doi.org/10.1016/S0168-1699(02)00100-X

Costa, N. R., Lourenço, J., & Pereira, Z. L. (2011). Desirability function approach: A review and performance evaluation in adverse conditions. *Chemometrics and Intelligent Laboratory Systems*, *107*(2), 234–244. https://doi.org/10.1016/j.chemolab.2011.04.004

D Bloice, M., Stocker, C., & Holzinger, A. (2017). Augmentor: An Image Augmentation Library for Machine Learning. *ArXiv Preprint ArXiv:1708.04680*. https://doi.org/10.21105/joss.00432

Dalal, N., & Triggs, B. (2005). Histograms of oriented gradients for human detection. *Computer Society Conference on Computer Vision and Pattern Recognition (CVPR'05)*. https://doi.org/10.1109/CVPR.2005.177

Dandois, J. P., & Ellis, E. C. (2013). High spatial resolution three-dimensional mapping of vegetation spectral dynamics using computer vision. *Remote Sensing of Environment*, *136*, 259–276. https://doi.org/10.1016/j.rse.2013.04.005

Dennis, J.E. Junior; Schnabel, R. B. (1996). *Numerical Methods for Unconstrained*

*Optimization and Nonlinear Equations* (U. ofWashington Robert E. O'Malle, Jr. (ed.); First). Society for Industrial and Applied Mathematics.

Fan, S., Li, J., Zhang, Y., Tian, X., Wang, Q., He, X., Zhang, C., & Huang, W. (2020). On line detection of defective apples using computer vision system combined with deep learning methods. *Journal of Food Engineering*, *286*, 110102. https://doi.org/10.1016/J.JFOODENG.2020.110102

Ge, Z., Liu, S., Wang, F., Li, Z., & Sun, J. (2021). YOLOX: Exceeding YOLO Series in 2021. *ArXiv Preprint ArXiv:2107.08430*, 1–7. http://arxiv.org/abs/2107.08430

Hardin, R. G., Huang, Y., & Poe, R. (2018). Detecting Plastic Trash in a Cotton Field With a UAV. *Beltwide Cotton Conferences*, 521–527. http://www.cotton.org/beltwide/proceedings/2005-2018/data/conferences/2018/papers/18538.pdf

Himmelsbach, D. S., Hellgeth, J. W., & McAlister, D. D. (2006). Development and use of an Attenuated Total Reflectance/Fourier Transform Infrared (ATR/FT-IR) spectral database to identify foreign matter in cotton. *Journal of Agricultural and Food Chemistry*, *54*(20), 7405–7412. https://doi.org/10.1021/jf052949g

Howard, A., Sandler, M., Chen, B., Wang, W., Chen, L. C., Tan, M., Chu, G., Vasudevan, V., Zhu, Y., Pang, R., Le, Q., & Adam, H. (2019). Searching for mobileNetV3. *Proceedings of the IEEE/CVF International Conference on Computer Vision*, 1314–1324. https://doi.org/10.1109/ICCV.2019.00140

Jocher, G., Stoken, A., Chaurasia, A., Borovec, J., NanoCode012, TaoXie, Kwon, Y., Michael, K., Changyu, L., Fang, J., V, A., Laughing, tkianai, yxNONG, Skalski, P., Hogan, A., Nadar, J., imyhxy, Mammana, L., … wanghaoyang0106. (2021). *ultralytics/yolov5: v6.0 -*

YOLOv5n "Nano" models, Roboflow integration, TensorFlow export, OpenCV DNN
support. https://doi.org/10.5281/ZENODO.5563715

Kalinke, T., Tzomakas, C., & Seelen, W. V. (1998). A texture-based object detection and an
adaptive model-based classification. *International Conference on Intelligent Vehicles*,
341–346. http://citeseerx.ist.psu.edu/viewdoc/summary?doi=10.1.1.33.526

Kist, A. M. (2021). *Glottis Analysis Tools - Deep Neural Networks. Zenodo*.
https://doi.org/10.5281/zenodo.4436985

Li, Tsung-Yi;Goyal, Priya;Girshick, Ross;He, Kaiming;Dollar, P. (2018). Focal Loss for Dense
Object Detection. *International Conference on Computer Vision*.

Lin, T. Y., Maire, M., Belongie, S., Hays, J., Perona, P., Ramanan, D., Dollár, P., & Zitnick, C.
L. (2014). Microsoft COCO: Common Objects in Context. *European Conference on
Computer Vision*, *8693 LNCS*(PART 5), 740–755. https://doi.org/10.1007/978-3-319-
10602-1_48

Liu, G., Nouaze, J. C., Touko Mbouembe, P. L., & Kim, J. H. (2020). YOLO-Tomato: A
Robust Algorithm for Tomato Detection Based on YOLOv3. *Sensors (Basel, Switzerland)*,
*20*(7), 1–20. https://doi.org/10.3390/s20072145

Liu, S., Qi, L., Qin, H., Shi, J., & Jia, J. (2018). Path Aggregation Network for Instance
Segmentation. *Computer Society Conference on Computer Vision and Pattern
Recognition*, 8759–8768. https://doi.org/10.1109/CVPR.2018.00913

Liu, X., Tang, G., & Zou, W. (2021). Improvement of Detection Accuracy of Aircraft in
Remote Sensing Images Based on YOLOV5 Model. *International Geoscience and Remote
Sensing Symposium IGARSS*, 4775–4778.
https://doi.org/10.1109/igarss47720.2021.9554234

Luque, A., Carrasco, A., Martín, A., & de las Heras, A. (2019). The impact of class imbalance in classification performance metrics based on the binary confusion matrix. *Pattern Recognition*, *91*, 216–231. https://doi.org/10.1016/j.patcog.2019.02.023

Mao, Q. C., Sun, H. M., Liu, Y. B., & Jia, R. S. (2019). Mini-YOLOv3: Real-Time Object Detector for Embedded Applications. *IEEE Access*, *7*, 133529–133538. https://doi.org/10.1109/ACCESS.2019.2941547

Obermiller, D. J. (2000). Multiple Response Optimization using JMP. *SAS Users Group International 22*. https://support.sas.com/resources/papers/proceedings/proceedings/sugi22/INFOVIS/PAPER178.PDF

Pelletier, M. G., Holt, G. A., & Wanjura, J. D. (2020). A Cotton Module Feeder Plastic Contamination Inspection System. In *AgriEngineering* (Vol. 2, Issue 2). https://doi.org/10.3390/agriengineering2020018

Raghavendra, U., Fujita, H., Bhandary, S. V., Gudigar, A., Tan, J. H., & Acharya, U. R. (2018). Deep convolution neural network for accurate diagnosis of glaucoma using digital fundus images. *Information Sciences*, *441*, 41–49. https://doi.org/10.1016/j.ins.2018.01.051

Redmon, J., & Farhadi, A. (2018). YOLOv3:An Incremental Improvement. *Computer Vision and Pattern Recognition*. https://pjreddie.com/media/files/papers/YOLOv3.pdf

Ren, X., & Ramanan, D. (2013). Histograms of Sparse Codes for Object Detection. *Computer Vision and Pattern Recognition*, 3246–3253. https://doi.org/10.1109/CVPR.2013.417

Robbins, R. F. (2018). New Extraneous Matter Code for Plastic Contaminants in Cotton Samples. In *USDA Agricultural Marketing Service-Cotton & Tobacco Program*. https://mymarketnews.ams.usda.gov/sites/default/files/resources/2020-02/New Extraneous

Matter Code for Plastic Contaminants in Cotton Samples-06292018.pdf

Ruder, S. (2016). An overview of gradient descent optimization algorithms. *ArXiv Preprint ArXiv:1609.04747*, 1–14. http://arxiv.org/abs/1609.04747

Samie, F., Tsoutsouras, V., Bauer, L., Xydis, S., Soudris, D., & Henkel, J. (2016). Computation offloading and resource allocation for low-power IoT edge devices. *3rd World Forum on Internet of Things (WF-IoT)*, 7–12. https://doi.org/10.1109/WF-IoT.2016.7845499

Solawetz, Jacob;Nelson, J. (2020). *How to Train YOLOv5 On a Custom Dataset*. https://blog.roboflow.com/how-to-train-yolov5-on-a-custom-dataset/

The SciPy Community. (2021). *Statistical functions (scipy.stats)*. https://docs.scipy.org/doc/scipy/reference/stats.html

Tzutalin. (2015). *LabelImg. Git code*. https://github.com/tzutalin/labelImg

USDA-Natural Resources Conservation Service. (2020). *Web Soil Survey*. https://websoilsurvey.sc.egov.usda.gov/App/HomePage.htm

Wanjura, J., Pelletier, M., Ward, J., Hardin, B., & Barnes, E. (2020). Prevention of Plastic Contamination When Handling Cotton Modules. *Cotton Incorporated*, *August*, 1–10.

Whitehill, J., Littlewort, G., Fasel, I., Bartlett, M., & Movellan, J. (2009). Toward practical smile detection. *IEEE Transactions on Pattern Analysis and Machine Intelligence*, *31*(11), 2106–2111. https://doi.org/10.1109/TPAMI.2009.42

Whitelock, D., Pelletier, M., Thomasson, A., Buser, M., Xu, B., Delhom, C., & Hardin, R. (2018). Current University and USDA Lab Cotton Contamination Research. *Beltwide Cotton Conferences*, *Figure 3*, 516–520. http://www.cotton.org/beltwide/proceedings/2005-2019/data/conferences/2018/papers/18518.pdf#page=1

Xie, S., & Braga-Neto, U. M. (2019). On the Bias of Precision Estimation Under Separate Sampling. *Cancer Informatics*, *18*. https://doi.org/10.1177/1176935119860822

Yadav, P. (2021). *image_splitter*. https://github.com/pappuyadav/image_splitter

Yadav, P. K., Thomasson, J. A., Hardin, R. G., Searcy, S. W., Braga-Neto, U. M., Popescu, S. C., Martin, D. E., Rodriguez, R., Meza, K., Enciso, J., Solorzano, J., & Wang, T. (2022a). *Volunteer cotton plant detection in corn field with deep learning*. *1211403*(June), 3. https://doi.org/10.1117/12.2623032

Yadav, P. K., Thomasson, J. A., Hardin, R., Searcy, S. W., Braga-neto, U., Popescu, S. C., Martin, D. E., Rodriguez, R., Meza, K., Enciso, J., Solorzano, J., & Wang, T. (2022b). Detecting Volunteer Cotton Plants in a Corn Field with Deep Learning on UAV Remote-Sensing Imagery. *ArXiv Preprint ArXiv:2207.06673*, 1–38. https://doi.org/10.48550/arXiv.2207.06673

Yadav, P. K., Thomasson, J. A., Searcy, S. W., Hardin, R. G., Braga-Neto, U., Popescu, S. C., Martin, D. E., Meza, K., Enciso, J., & Diaz, J. S. (2022c). Assessing the Performance of Yolov5 Algorithm for Detecting Volunteer Cotton Plants in Corn Fields at Three Different Growth Stages. *Artificial Intelligence in Agriculture*, *6*. https://doi.org/10.2139/ssrn.4188683

Yadav, P. K., White, E. L., Thomasson, J. A., Cholula, U., Marconi, T., & Enciso, J. (2020). Application of UAV Remote Sensing for Detecting Plastic Contaminants in Cotton Fields. *Beltwide Cotton Conferences*.

Yan, B., Fan, P., Lei, X., Liu, Z., & Yang, F. (2021). A real-time apple targets detection method for picking robot based on improved YOLOv5. *Remote Sensing*, *13*(9), 1–23. https://doi.org/10.3390/rs13091619

Zhang, Y., Zhang, W., Yu, J., He, L., Chen, J., & He, Y. (2022). Complete and accurate holly

fruits counting using YOLOX object detection. *Computers and Electronics in Agriculture*,

*198*(January), 107062. https://doi.org/10.1016/j.compag.2022.107062

Zhou, F., Zhao, H., & Nie, Z. (2021). Safety Helmet Detection Based on YOLOv5.

*International Conference on Power Electronics, Computer Applications, ICPECA 2021*,

6–11. https://doi.org/10.1109/ICPECA51329.2021.9362711